\documentclass[conference]{IEEEtran}
\IEEEoverridecommandlockouts
\usepackage{cite}
\usepackage{amsmath,amssymb,amsfonts}
\usepackage{graphicx}
\usepackage{textcomp}
\usepackage{xcolor}

\def\BibTeX{{\rm B\kern-.05em{\sc i\kern-.025em b}\kern-.08em
    T\kern-.1667em\lower.7ex\hbox{E}\kern-.125emX}}
\usepackage{subcaption}
\usepackage[]{algorithm}
\usepackage{algpseudocode}
\usepackage{listings}
\usepackage{tabularx}

\begin{document}

\title{HW-Aware Initialization of DNN Auto-Tuning to Improve Exploration Time and Robustness}

\author{
\IEEEauthorblockN{Dennis Rieber}
\IEEEauthorblockA{\textit{Corporate Research} \\ \textit{Robert Bosch GmbH}\\ Renningen, Germany \\ DennisSebastian.Rieber@de.bosch.com}
\and
\IEEEauthorblockN{ Moritz Reiber \hfill Oliver Bringmann}
\IEEEauthorblockA{\textit{Department of Embedded Systems} \\ \textit{University of T\"ubingen}\\ T\"ubingen, Germany \\	moritz.reiber@uni-tuebingen.de \\ oliver.bringmann@uni-tuebingen.de}
\and
\IEEEauthorblockN{Holger Fr\"oning}
\IEEEauthorblockA{\textit{Computing Systems Group} \\ \textit{Heidelberg University}\\ Heidelberg, Germany\\ holger.froening@ziti.uni-heidelberg.de}
}
\maketitle

\begin{abstract}
The process of optimizing the latency of DNN operators with ML models and hardware-in-the-loop, called auto-tuning,  has established itself as a pervasive method for the deployment of neural networks. From a search space of loop-optimizations, the candidate providing the best performance has to be selected. Performance of individual configurations is evaluated through hardware measurements. The combinatorial explosion of possible configurations, together with the cost of hardware evaluation makes exhaustive explorations of the search space infeasible in practice. Machine Learning methods, like random forests or reinforcement learning are used to aid in the selection of candidates for hardware evaluation.
For general purpose hardware like x86 and GPGPU architectures impressive performance gains can be achieved, compared to hand-optimized libraries like cuDNN.
The method is also useful in the space of hardware accelerators with less wide-spread adoption, where a high-performance library is not always available.
However, hardware accelerators are often less flexible with respect to their programming which leads to operator configurations not executable on the hardware target. This work evaluates how these invalid configurations affect the auto-tuning process and its underlying performance prediction model for the VTA hardware.
From these results, a validity-driven initialization method for AutoTVM is developed, only requiring $41.6\%$ of the necessary hardware measurements to find the best solution, while improving search robustness.
% Motivation - Ever growing space of deep learnign accelerators and operators
% Problem - Everything is still one manually. Challenges for automation are search space and embedding
% Approach - Search space and pattern matching in graph-based IR
% Result - 
% Conclusion -

\end{abstract}


\section{Introduction}
High performance implementations of DNN operators, regardless of the specific target metric, are often enabled by manually implemented libraries, written in low level languages like C or assembly for a specific hardware target~\cite{Chetlur2014cuDNNEP}. They aim to optimally utilize the architecture and memory hierarchy of the system by manipulating the loops of the original program. In essence, the load, store and compute operations of a specific operator are reordered in such a way, that cache misses, bank conflicts, unnecessary memory accesses, pipeline stalls, register spills or other performance bugs are avoided, or at least minimized. Such implementations are time consuming to write and require a deep understanding of the targeted system and workload. Hardware-in-the-loop optimization, or auto-tuning, helps to automate this process.
Auto-tuning probes a search space of different implementations for the operator. This space is generated from the combination of different, individual loop optimizations. The search-spaces size often prevents exhaustive attempts for any meaningful problem.

Auto-tuning is especially interesting for hardware accelerators, which often are not ubiquitous enough to justify extensive library development. At the same time they have tight resource budgets and more rigid code execution mechanisms, making them hard to program for. For example, on the \textit{Versatile Tensor Accelerator} (VTA)\cite{VTA} it is not possible to define a tiling scheme that would overflow one of the local buffers. While a CPU would be less efficient with such code, the VTA software pipeline refuses to compile such code entirely. Many parameter configurations can cause this issue, as well as different operator instances or deployment strategies.
A consequence is a large subspace of \textit{invalid configurations} in the auto-tuning search space. Many auto-tuning tools, like AutoTVM ~\cite{AutoTVM:10.5555/3327144.3327258}, have at their core a statistical model to aid in the search by predicting the performance of individual configurations. However, when invalid configurations are a part of the search space, the model has to the balance the ability to discern valid from invalid configurations and accurately predict the performance at the same time. It is not well known how the models used in auto-tuning behave in such a context. Therefore, investigating the effects of invalid configurations on the auto-tuning process is of interest. 
This work presents the following contributions:
\begin{itemize}
	\item An analysis of occurrence and effect of such invalid configurations in a search space for Conv2D on the VTA accelerator. First, the distribution in different search spaces is analysed. Then, how invalid configurations influence the performance prediction model at the core of AutoTVM~\cite{AutoTVM:10.5555/3327144.3327258}, as well as the auto-tuning process itself.
	\item From these findings, a procedure to prevent the negative influence of the invalid configurations on the auto-tuning process is developed and evaluated.
\end{itemize}

%Section \ref{c5:relatedwork} introduces auto-tuning with AutoTVM and explores related publications.
In Section \ref{c5-sec:searchspaceanalysis}, the  occurrence, distribution and impact of invalid configurations is explored.
This exploration is the basis for possible improvements, which are described 
in Section \ref{c5:methods}. Evaluation of these improvement over a series of experiments is detailed in Section \ref{c5-sec:results}, where reduced tuning times and higher robustness are demonstrated.

\section{AutoTVM Background}\label{c5-sec:background}
An auto-tuning search space is the Cartesian product of all different possible optimization parameters, like loop tiling, reordering and threading strategies, as shown by Table \ref{tab:knobs}. A configuration is a vector of one possible value from each parameter. AutoTVM works by selecting batches of configurations to evaluate in hardware in each tuning epoch. To select them, the tuner optimizes the ranking loss over $n$ configurations via simulated annealing (SA), predicted by a statistical model. These top-$n$ candidates are selected for evaluation on hardware. The measurement results are then used to further train the statistical model predicting the performance. AutoTVM uses gradient boosted trees~\cite{friedman2001gbt} as the underlying statistical method. 
The candidate vectors can be used as features directly (\textit{knob features}) or used to generate a feature set portable between operators (\textit{context relation features}). The latter also enables the training of the performance model on data from previous tuning runs, before the optimization of a new workload starts.

\begin{table}[t]
	\centering
	\caption[Knob example]{Example of tuning knobs for a convolution layer. 
		
	}
	
	\begin{tabular}{llr}
		\hline
		Name & Type & Possible values \\
		\hline
		[h,oc]\_threading & OtherOption & 1, 2 \\
		loop\_order  & OtherOption & 1, 2, 3, 4 \\
		tile\_b     & Split       & 1, 2 \\
		tile\_[h,w] & Split       & 1, 2, 4, 7, 8, 14, 16, 28, 56, 112 \\
		tile\_ci    & Split       & 1, 2, 4  \\
		tile\_co    & Split       & 1, 2, 4  \\
		\hline
	\end{tabular}
	\label{tab:knobs}
\end{table}

\section{VTA Auto-Tuning Search Space Analysis}\label{c5-sec:searchspaceanalysis}
When invalid candidates are in the training data for the performance prediction model inside the auto-tuner, multiple questions arise. Mainly, how to present these configurations towards the model and subsequently,  how does this representation influence the model?
Two scenarios are possible for representation. The first is to ignore invalid candidates and only use valid performance results. This is useful to avoid a distortion of the model and focusses on the prediction task. However, this poses the question, if it also makes the model blind to potential invalid configurations.
AutoTVM implements a second option, where invalid configurations are fed back into the model with a heavy performance penalty. The intention is to train the model towards avoiding the invalid configurations. However, it is not known how the model reacts to this additional complexity. Can the model discern valid configurations and predict performance well enough?

\begin{table*}[t]
	\small
	\caption[Workload details]{Workloads set: The workload ID is the respective position in the 
		Baidu DeepBench benchmark suite.}
	\centering
	\begin{tabular}{rrrrrrrrr r}
		\hline
		Workload ID & batch & channel out & image & kernel & channel in & stride & pad & search space size & valid ratio\\
		\hline
		3  &  1 &   32  &  $79 \times 341$ &  $10 \times 5$ &   32  &  [2 2]  &  [0 0] & 768 &0.060\\ 
		5  &  4 &   32  &  $79 \times 341$ &  $10 \times 5$ &   32  &  [2 2]  &  [0 0] & 3072&0.068\\ 
		8  &  1 &   64  &  $12 \times 120$ &  $ 3 \times 3$ &   32  &  [1 1]  &  [1 1] & 9216&0.067\\ 
		17  &  1 &  256  &  $56 \times  56$ &  $ 3 \times 3$ &  128  &  [1 1]  &  [1 1] & 20480&0.027\\ 
		42  &  1 &   64  &  $56 \times  56$ &  $ 3 \times 3$ &   64  &  [1 1]  &  [1 1] & 9216&0.047\\ 
		48  &  1 & 1024  &  $14 \times  14$ &  $ 1 \times 1$ &  256  &  [2 2]  &  [0 0] & 2240&0.151\\ 
		53  &  2 &  128  &  $28 \times  28$ &  $ 3 \times 3$ &  128  &  [1 1]  &  [1 1] & 18433&0.035\\ 
		59  &  2 &  512  &   $7 \times   7$ &  $ 1 \times 1$ & 2048  &  [2 2]  &  [3 3] & 6145&0.008\\ 
		76  &  1 &   64  & $112 \times 112$ &  $ 1 \times 1$ &   64  &  [1 1]  &  [0 0] & 14400&0.088\\ 
		78  &  1 &   64  &  $56 \times  56$ &  $ 1 \times 1$ &  256  &  [1 1]  &  [0 0] & 15361&0.099\\ 
		92  &  2 &   64  & $112 \times 112$ &  $ 1 \times 1$ &   64  &  [1 1]  &  [0 0] & 28800&0.082\\ 
		106  &  2 & 2048  &  $14 \times  14$ &  $ 1 \times 1$ & 1024  &  [2 2]  &  [0 0] & 7169&0.122\\ 
		107  &  2 &  512  &   $7 \times   7$ &  $ 1 \times 1$ & 2048  &  [1 1]  &  [0 0] & 6144&0.231\\ 
		\hline
	\end{tabular}
	
	\label{c5-tab:workload_props}
\end{table*}

To analyse this impact, a subset of the Baidu DeepBench Inference Benchmark\footnote{{https://github.com/baidu-research/DeepBench}, accessed 05.2022} was picked for further experiments. The selected convolutions are presented in Table \ref{c5-tab:workload_props}. The ID in the table refers to the ID in the benchmark suite. The original auto-tuning template was modified to include additional loop permutations, which increase the search space. These changes create both, valid and invalid configurations. For these workloads, an exhaustive exploration of the search spaces was performed. This dataset serves as baseline for the analysis in Section~\ref{c5-ssec:lack_of_robustness} and the experiments in Section~\ref{c5-sec:results}. All experiments rely on the \textit{knob} features. They represent the selected configuration of loop tiling, permutation and parallelization factors of the specific candidate in a flattened vector.

The right-most column in Table \ref{c5-tab:workload_props} presents the share of valid configurations of each search space. For all workloads, a majority of configurations is invalid. When $v$ is the amount of valid configurations and $t$ is the total search space size, defined as the $ratio = \frac{v}{t}$.

The mean ratio over all workloads is $0.083$, with $0.008$ and $0.231$ being the minimum and maximum, respectively.

\subsection{Spatial Organization of Valid Configurations}\label{sec:graph_clusters}
Combining search-space knobs forms a regular, cartesian grid. When interpreting this grid as a graph, it is possible to analyse the spatial distribution of valid and invalid configurations in the search spaces. To interpret the search space as a graph, two nodes are connected if the Manhattan distance~\cite{krause1986taxicab} between their indices in the grid is exactly one.

\begin{figure}[t]
	\centering
	\includegraphics[height=.14\textheight]{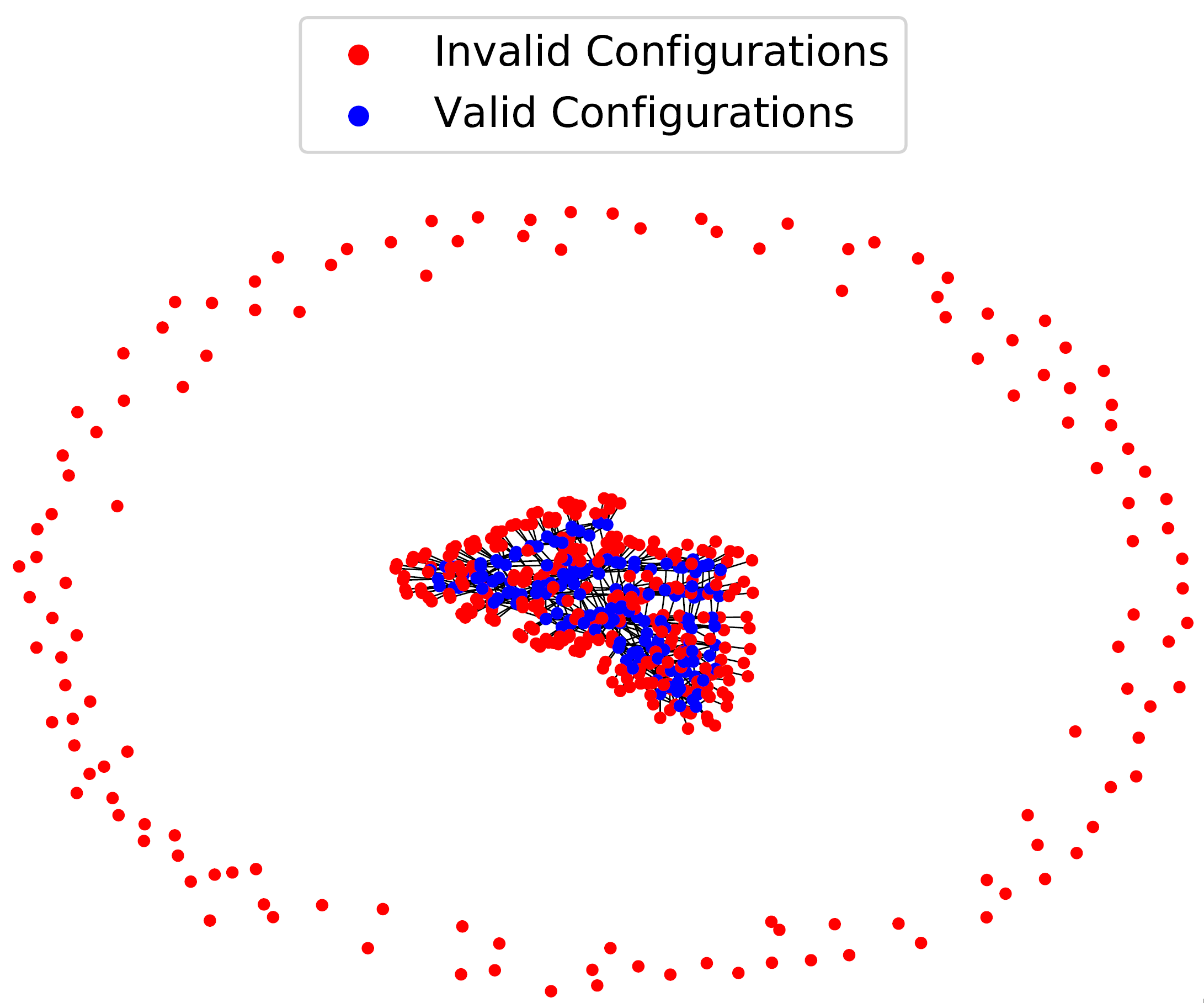}
	
	\caption[Search space in graph representation]{Visualization of layer 48 search space as a graph with connections between valid candidates. }
	\label{c5-fig:graph_clus_0}
\end{figure}

Only drawing edges connected to a valid candidate yields the graph Figure in \ref{c5-fig:graph_clus_0}. This graph shows a clustering of valid configurations, indicating a spatial relationship in the search space. This behaviour is not unique, as it occurred in all search spaces of Table \ref{c5-tab:workload_props}. This observation is valuable, as it can be leveraged to improve the optimization process. However, a detailed analysis of the effects forming clusters is not deemed helpful, as every workload configuration creates invalid configurations for many reasons.

\subsection{Standalone Model Evaluation}
%The AutoTVM model learns to rank the throughput in $GFLOp/s$ of different candidates, invalid candidates are scored with $0.0\ GFLOp/s$. 
As AutoTVM adds invalid configurations with a penalty to the performance model training set, it poses the question, how well the model can handle this. Specifically, if valid but low performing and invalid configurations are related in the search space and if treating them similarly distorts the prediction ability of the model.
The following sections present the used methods to quantify the effects of invalid configurations on the ranking performance.

\subsubsection{Accuracy of Pairwise Comparisons}
Here, $p_i$ is the predicted performance and $m_i$ the measured performance. Invalid configurations
are scored with $m_i = 0$. The accuracy of pairwise
comparisons is then defined as
\begin{equation}
\small
\operatorname{accuracy} = \frac{1}{n^2 - \frac{n}{2}} \sum_{i>j} \begin{cases}
&1 \text{ if sign}(m_i - m_j) = \text{sign}(p_i - p_j) \\
&0 \text{ else}  
\end{cases}    \
\end{equation}
or the number of correct comparisons divided by the total number of comparisons.
Comparisons between pairs of invalid configurations are ignored when computing the total accuracy, as there is no well-defined ranking among them.

% Precision
\subsubsection{Precision}
The share of valid configurations among all 
configurations in the ranking is the \textit{precision} of a model. 
Since AutoTVM selects the top-$n$ candidates from the
total ranking, the $precision@n$  is the ratio of valids in these $n$ configurations. It is defined as 

\begin{equation}
\small
\operatorname{precision}@n = \frac{1}{n}\sum_{i = 1}^{n} \begin{cases}
&1 \text{ if } \operatorname{valid}(x_{r_i}) \\
&0 \text{ else}
\end{cases},
\end{equation} with $x_{r_i}$ being a specific configuration from the search space at place $i$ in the ranking.
% NDCG
\subsubsection{Normalized Discounted Cumulative Gain}
Precision is only a metric for the ability of the model to retrieve valid configurations. It does not give indication of ranking quality in the top-$n$ candidates. The normalized discounted cumulative gain $nDCG$ is a metric to quantify this ranking ability.

Discounted cumulative gain (DCG) is defined as 

\begin{equation}
\operatorname{DCG}@n = \sum_{i=1}^n \frac{c_{r_i}}{\log_2(r_i + 1)},
\end{equation}

the sum of true ranking scores $c_{r_i}$, discounted by their respective position $r_i$ produced in the ranking.
The nDCG is computed by dividing DCG with the \textit{ideal} discounted cumulative gain (iDCG),
the maximum DCG score, achieved by an optimal ranking $r^*$ 

\begin{equation}
\operatorname{iDCG}@n = \sum_{i=1}^n \frac{c_{r^*_i}}{\log_2(r^*_i + 1)}
\end{equation}

Thus, the normalized discounted cumulative gain is computed by

\begin{equation}
\operatorname{nDCG}@n = \frac{DCG@n}{iDCG@n}
\end{equation}

Each time a candidate among the top-$n$ is ranked above another candidate with better measured performance, the score decreases. This quantifies the ranking quality. And since invalid configurations in the top-$n$ will always have a worse measured performance than any valid configuration, the metric can also be used to measure ability of the model to discerning valid and invalid configurations.

% Experiments & Results
\subsubsection{Experiments \& Results}

\begin{figure*}[h]
	\centering
	\begin{subfigure}{0.24\textwidth}
		\centering
		\includegraphics[width=\textwidth]{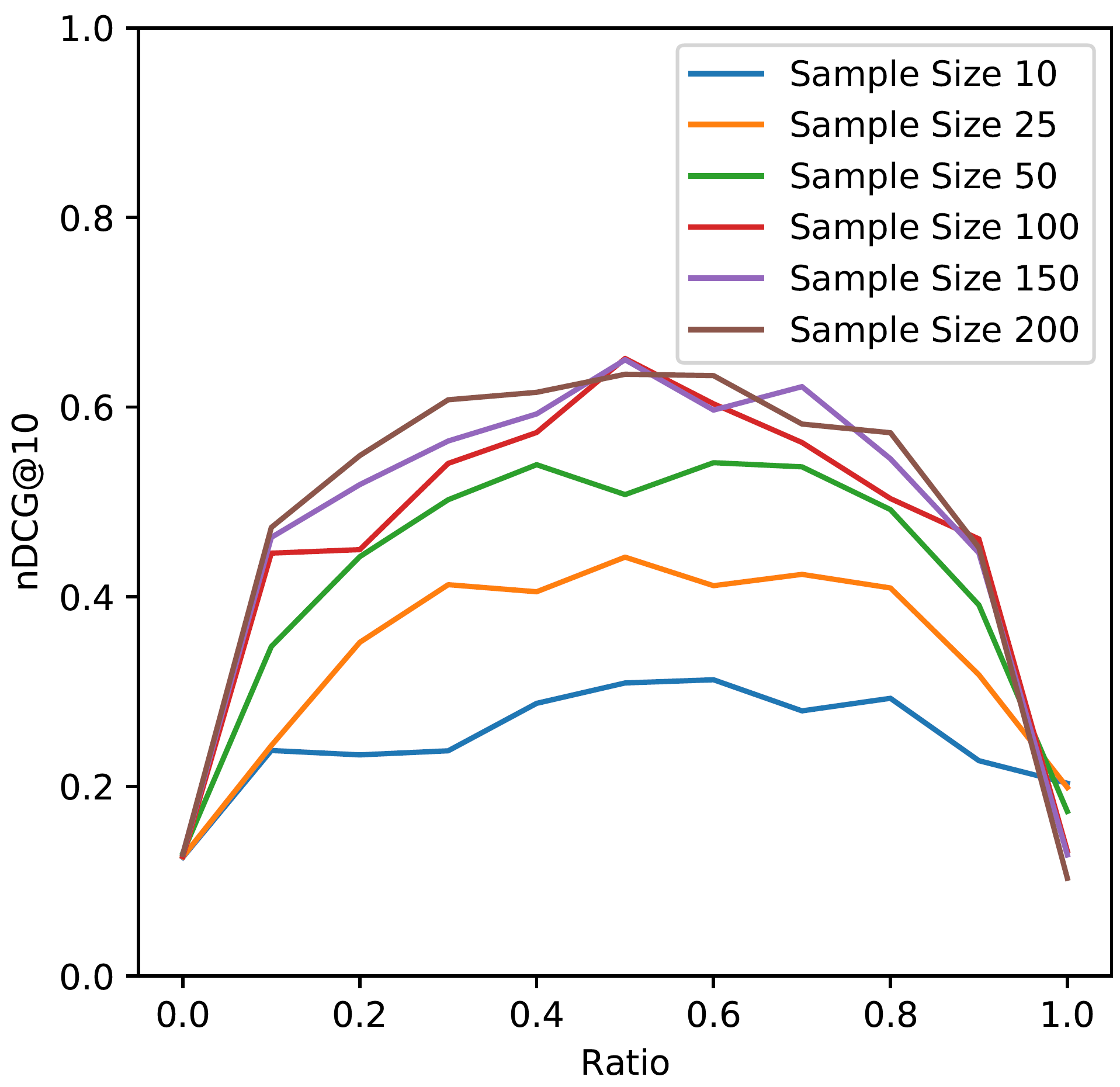}
		\caption{{nDCG}}
		\label{fig:cm_ndcg}
	\end{subfigure}
	\begin{subfigure}{0.24\textwidth}
		\centering
		\includegraphics[width=\textwidth]{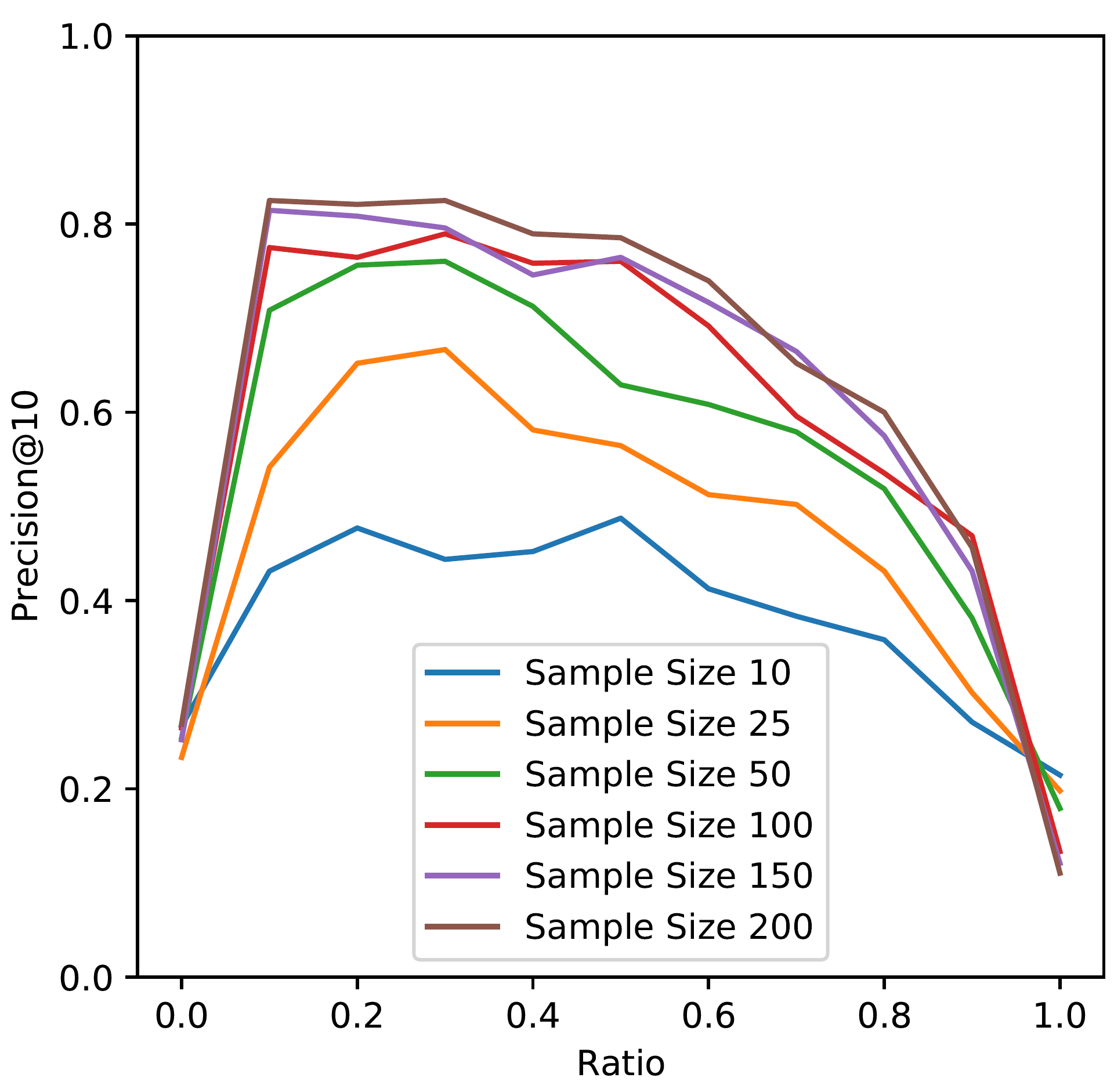}
		\caption{Precision}
		\label{fig:cm_precision}
	\end{subfigure}    
	\begin{subfigure}{0.24\textwidth}
		\centering
		\includegraphics[width=\textwidth]{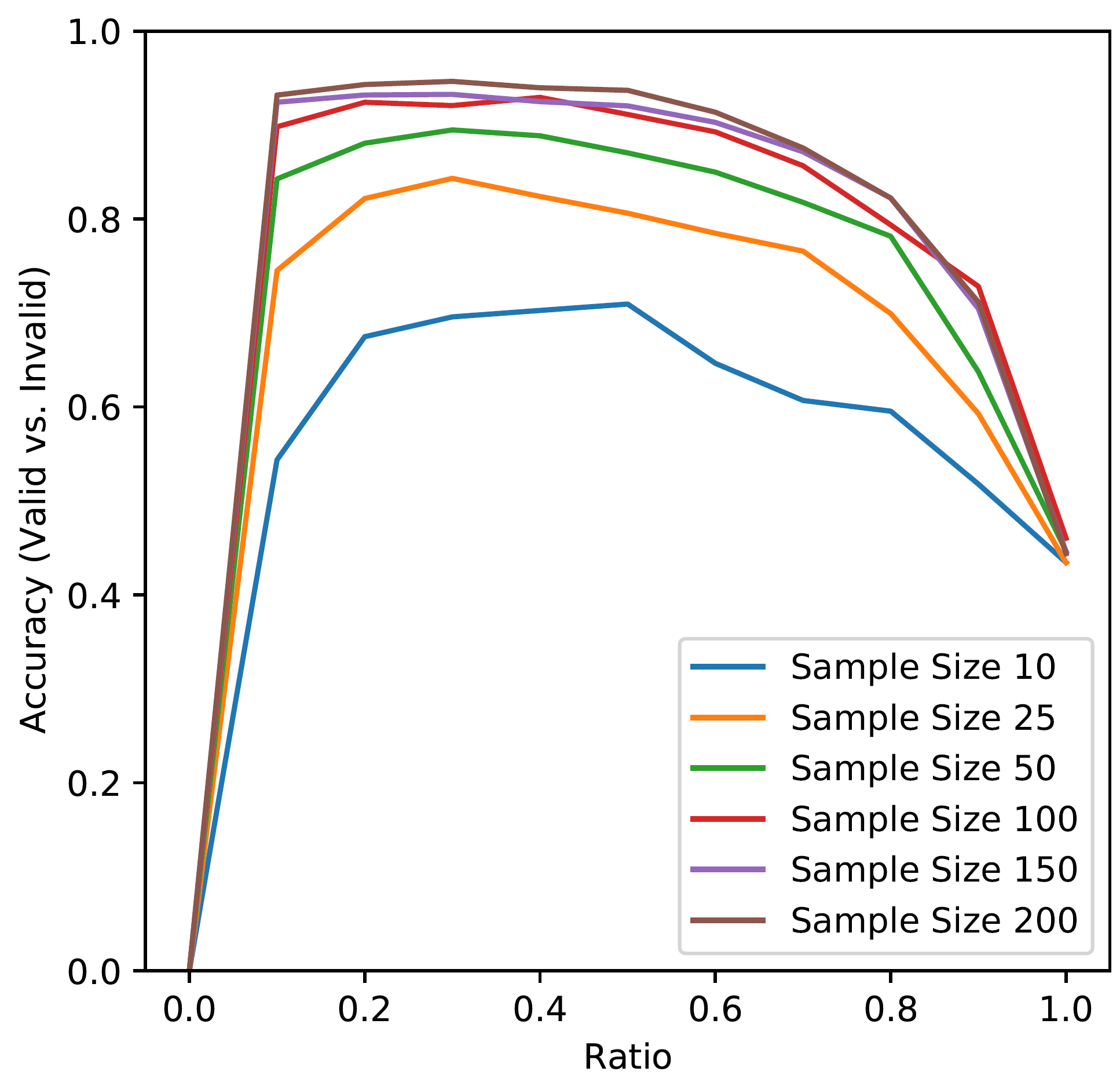}
		\caption{Accuracy Valid-Invalid}
		\label{fig:cm_accvi}
	\end{subfigure}
	\hfill
	\begin{subfigure}{0.24\textwidth}
		\centering
		\includegraphics[width=\textwidth]{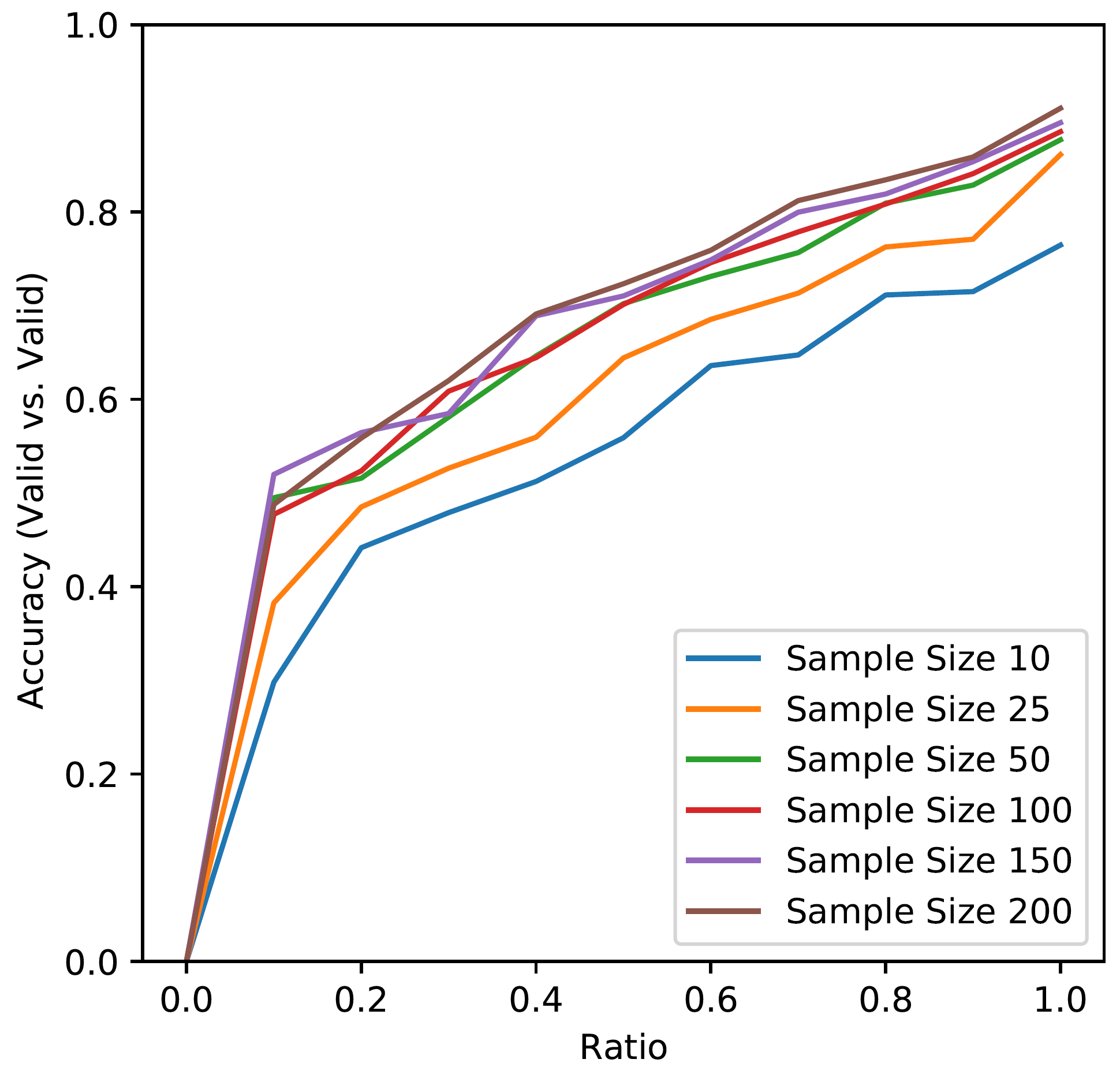}
		\caption{Accuracy Valid-Valid}
		\label{fig:cm_accvv}
	\end{subfigure}
	\caption{Cost-model performance evaluation.}
	\label{c5-fig:costmodel_ratio}
\end{figure*}

For the following experiments, workloads Baidu DeepBench Workloads $3-57$ \footnote{except workloads 6, 10, 15, 21, 27, 34 and 41, which are not executable on VTA} were randomly sampled for configurations and their performance measured on hardware. Based on this dataset, the model was evaluated in isolation. The configurations from every workload were randomly split into training data (75\%) and testing data (25\%).
Experiments for the presented metrics were done with a controlled ratio of valid configurations in the training set. A ratio of $0.3$ means that $30$ out of $100$ samples in the training are valid. This set was then used to train the performance model. This model's behaviour is then evaluated on an unmodified test set. 
Repeating this process for every workload and averaging over the results yielded the data in Figure \ref{c5-fig:costmodel_ratio}. 

Training the model on more data (larger sample size) improved the performance through all experiments. Experiments over different training batch sizes showed diminishing returns after 100 samples. This shows that the model requires relatively few samples to perform well. A low ratio of valid samples in the training set improves the models ability to distinguish between valid and invalid configurations, especially in the accuracy of valid-invalid (Figure \ref{fig:cm_accvi}) and precision (Figure  \ref{fig:cm_precision}) metrics.
With a higher frequency of valids, both precision (Figure  \ref{fig:cm_precision}) and valid-invalid (Figure  \ref{fig:cm_accvi}) performance drop significantly. When the model learns less about invalid configurations, it becomes harder for it to distinguish them from valid configurations. Even further, the diminishing ability to separate valids from invalids outweighs the theoretically better ranking performance in models trained on higher ratios.
Both observations are combined in Figure \ref{fig:cm_ndcg}, where across the full ratio scale a bell-shaped curve appeared. Since nDCG ranks both the retrieval and ranking ability, this makes sense. With a low ratio, the model fails at ranking the valid samples. This improves until an equilibrium at a ratio of 0.5 is reached. A further increase in valid samples then leads to a dropping score, as the ability to filter the invalids in the ranking correctly is reduced. Note that the score of nDCG should not be interpreted in same way as precision or accuracy. A score of 0.9 does not mean, that 90\% of decisions were correct. It rather allows relative comparisons in ranking ability between two nDCG scores, where a higher score means a better overall performance.

The observations confirm the hypothesis that ignoring invalid configurations distorts the performance model. When treating invalid configurations the same way as low performance configurations, the model's ranking quality is improved. Further, it seems that identifying invalid candidates has more influence on the ranking than the ability to rank among valids (see Figures \ref{fig:cm_accvv} and \ref{fig:cm_accvi}). Interesting is the equilibrium reached in the areas of balanced valid/invalid ratios, as the models seems to reach a sweet spot that, given enough data, has satisfactory ranking ability as well as the ability to distinguish between valid and invalid configurations.

\subsection{Tuning Robustness}\label{c5-ssec:lack_of_robustness}

\begin{figure}[t]
	\centering
	
	% \resizebox*{\textwidth}{!}{%
	\includegraphics[width=\linewidth]{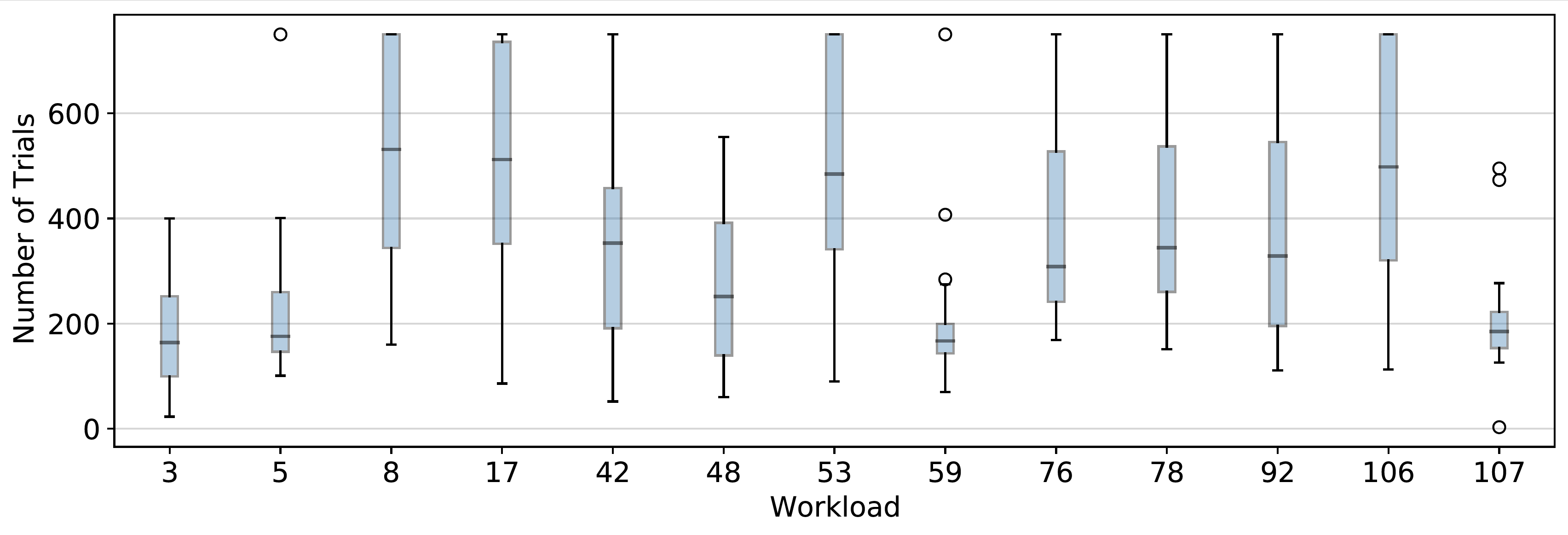}
	% }
	\caption[Number of trials: Baseline]{
		Number of trials to find the best candidate. Plotted over 20 runs with the auto-tuner, for each workload.}
	\label{c5-fig:num_trials_baseline}
\end{figure}

\begin{figure*}[t]
	\centering
	%	\begin{subfigure}[b]{0.4\textwidth}
	%		\centering
	%		\includegraphics[width=\textwidth]{img/robustness_53.png}
	%		\caption{Layer 53}
	%		\label{c5-fig:inconsistency_53}
	%	\end{subfigure}
	\begin{subfigure}{0.28\textwidth}
		\centering
		\includegraphics[width=\textwidth]{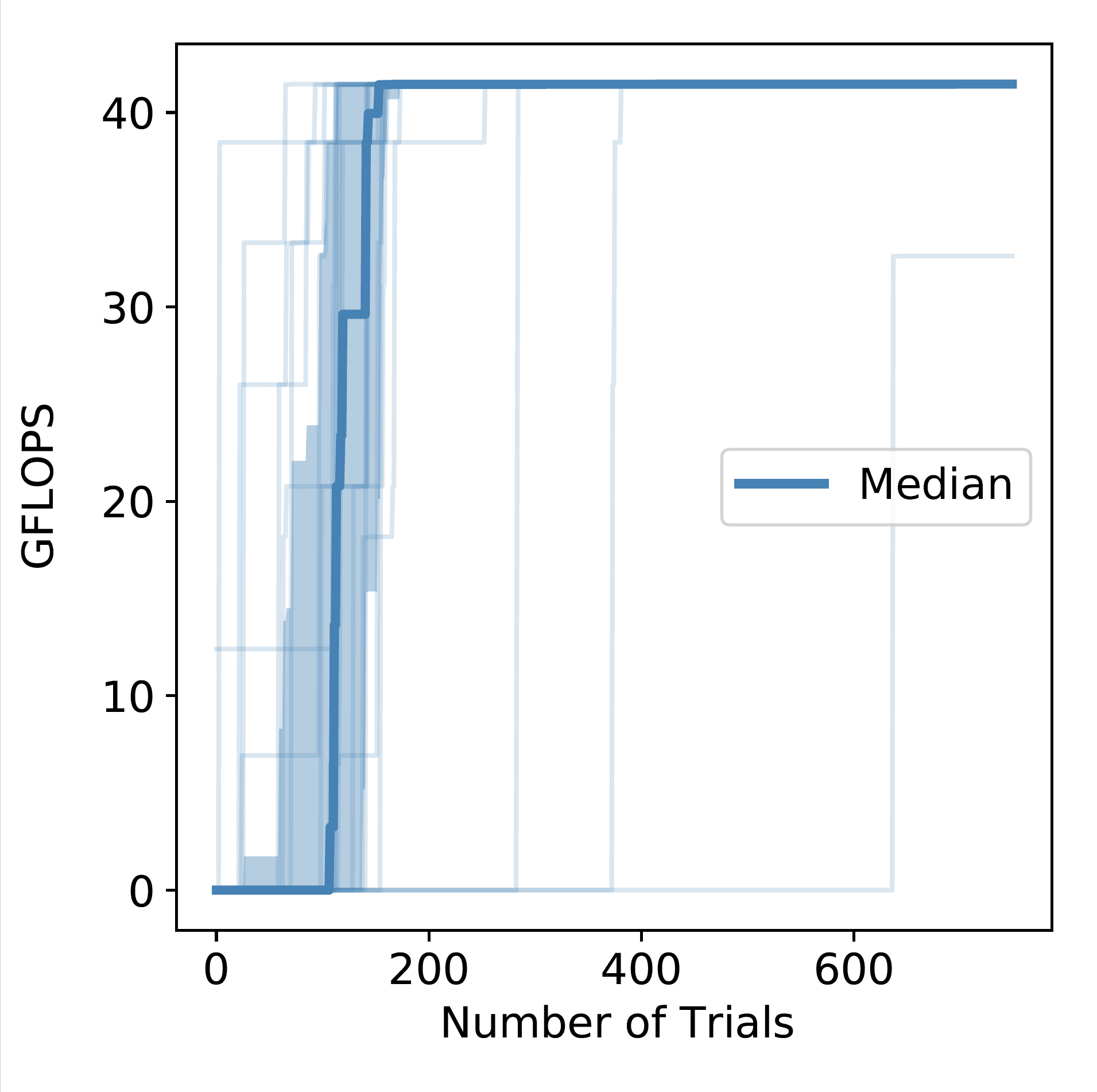}
		\caption{Layer 59}
		\label{c5-fig:inconsistency_59}
	\end{subfigure}
\hfill
	\begin{subfigure}{0.28\textwidth}
		\centering
		\includegraphics[width=\textwidth]{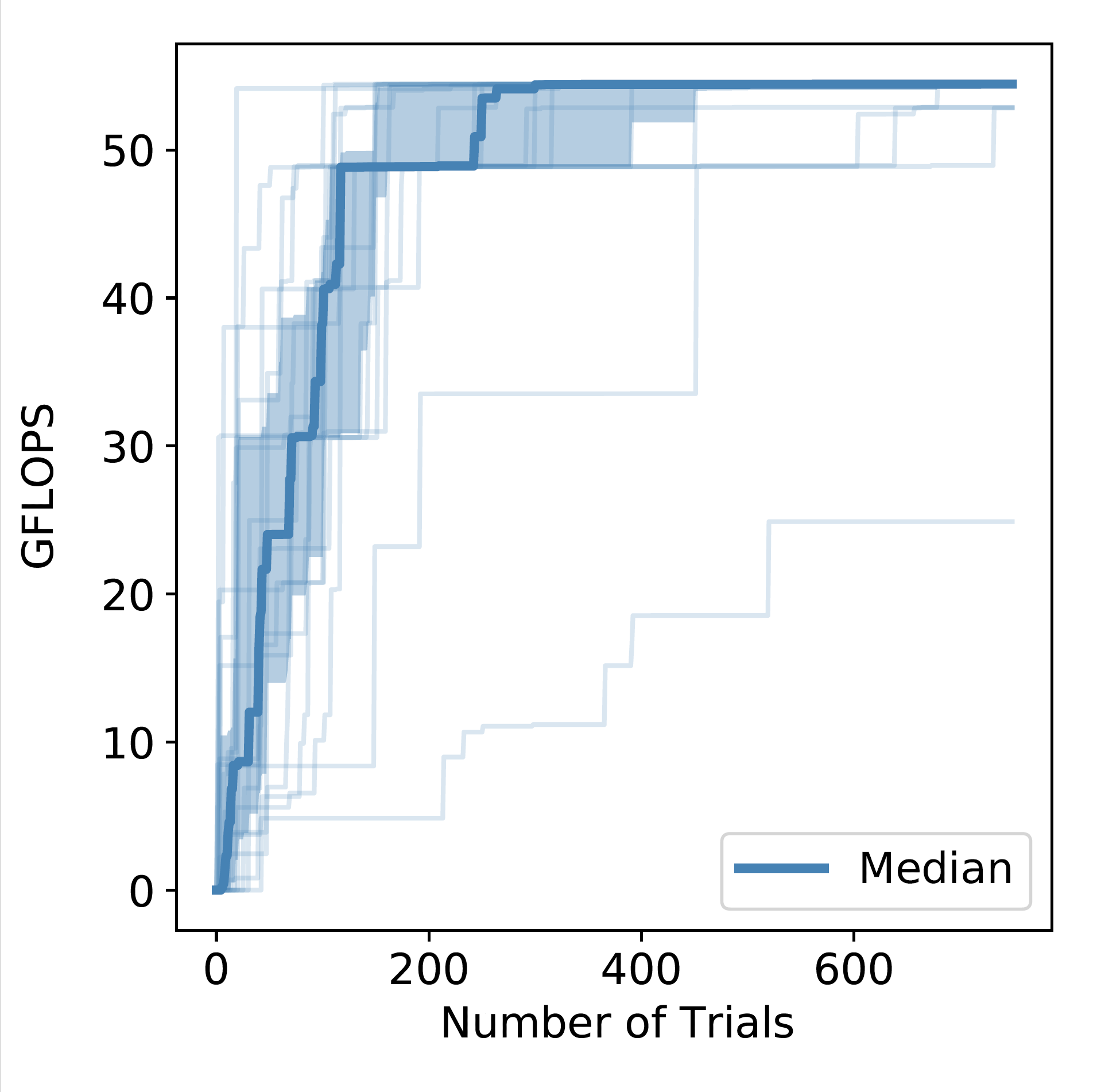}
		\caption{Layer 92}
		\label{c5-fig:inconsistency_92}
	\end{subfigure}
\hfill
	\begin{subfigure}{0.28\textwidth}
		\centering
		\includegraphics[width=\textwidth]{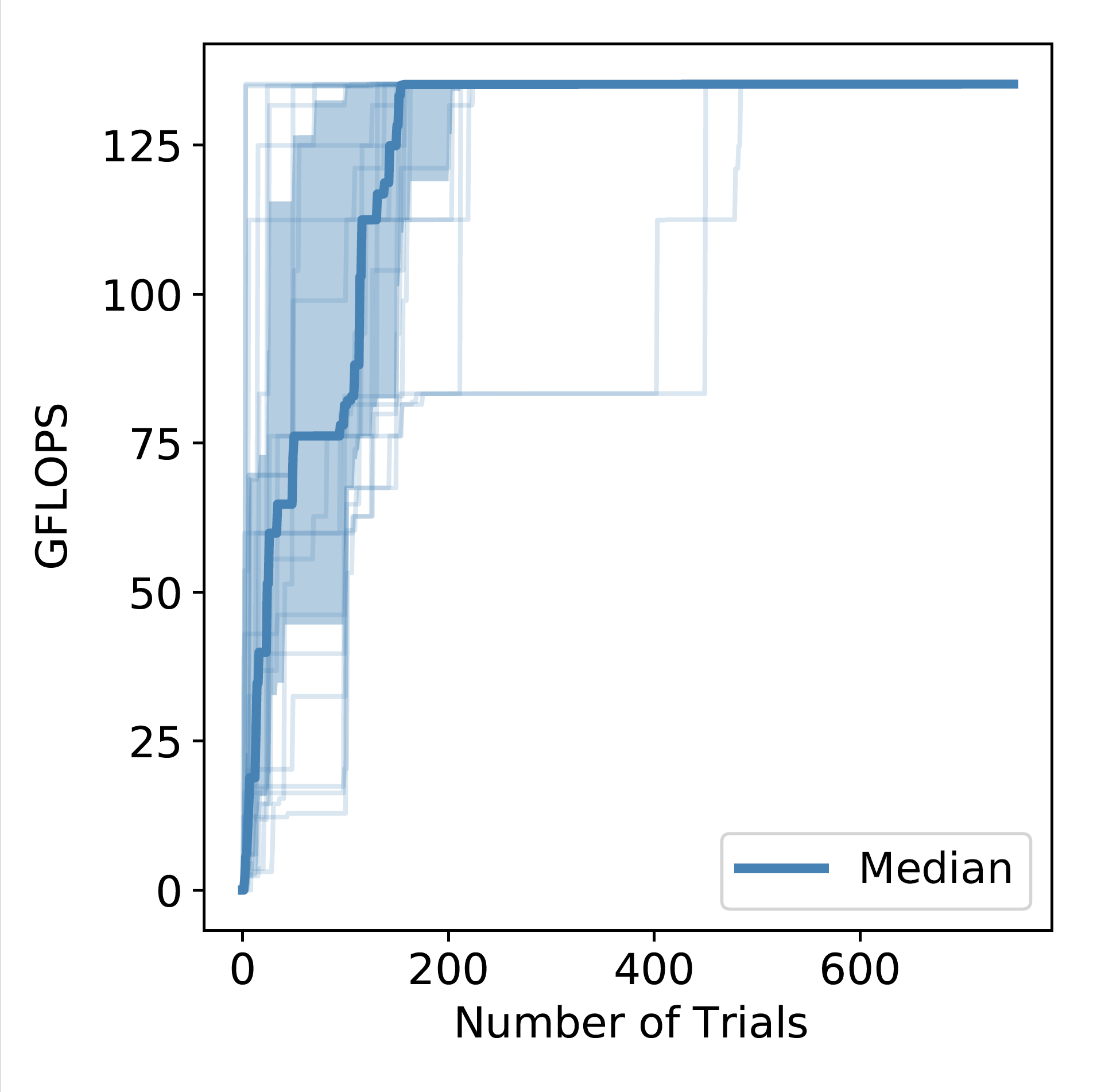}
		\caption{Layer 107}
		\label{c5-fig:inconsistency_107}
	\end{subfigure}
	\caption[Visualization of multiple autotuning runs]{Progression of multiple auto-tuning runs: Each curve shows the best performance (GFLOP/s) found in tuning run so far. The median over all runs is plotted bold and the shaded area is the inter-quartile range.}
	\label{c5-fig:inconsistency}
\end{figure*}

After evaluating the performance model in isolation, this section takes a look at the full auto-tuning process. The experiments use an untrained performance model, as is the default in AutoTVM. Tuning starts with a randomly sampled batch of size $e = 50$ and a total tuning duration of $t = 750$ trials. The optimization was repeated 20 times for all workloads in Table \ref{c5-tab:workload_props}.

The \textit{performance} of the auto-tuner is defined as the ability to find good configurations fast and consistently.
The following experiments use the median and interquartile range (IQR) as a measure for the former and latter criterion, respectively. The median gives a good indication on the \textit{convergence} time of the auto-tuner. IQR on the other hand describes central half of all median values in multiple experiments and is useful to determine how \textit{robust} the overall process is.

Figure \ref{c5-fig:num_trials_baseline} reports the amount of trials necessary to find the best performing configuration, aggregated over all 20 runs for each layer.
The first, and most important observation is the wide IQR across most workloads. 

For most layers, the tuner's convergence time varies greatly, hinting at large inconsistencies in the performance model's quality.  Figure \ref{c5-fig:inconsistency} shows all runs for layers 59, 92 and 107 which offers additional information on the behaviour of individual runs. For example, in layer 59 and 92 some runs did not find the best configuration during the 750 trials. This is shown by the traces of individual runs in Figure \ref{c5-fig:inconsistency}, that do not reach the upper performance bound. It also shows that the initial set of 50 randomly selected configurations greatly influence how long it takes the tuner to converge towards a good solution. Bad initial configurations can lead to plateaus in the search, which can last over multiple epochs. It is suspected, that the improvements found after several hundred trials can be attributed to random components in the search, rather than model quality.

\section{Validity Driven Model Initialization}\label{c5:methods}
Following the data in Section \ref{c5-sec:searchspaceanalysis}, a better initialization process for the performance model seems like a promising path to improve robustness and reduce the number of necessary hardware measurements. The main problem to address is the uniform sampling over a non-uniformly distributed solution space, as shown by the locality analysis in Figure \ref{c5-fig:graph_clus_0}.
The new process is aimed to improve the substantial robustness issues of auto-tuning runs by training a fresh model with a better initial data set, to achieve a balance in valid/invalid classification as well as performance prediction. Data in Figures \ref{fig:cm_ndcg} , \ref{fig:cm_precision} and \ref{fig:cm_accvi} shows that this is achievable with a balanced data set and relatively few samples for the model, starting from 50 to 100 training examples. 
Sampling valid candidates from the search space is the first step in this process.

\subsection{Neighbourhood-driven presampling}\label{c5-sec:presampling}
Instead of randomly selecting the initial measurement batch, a balanced set of valid and invalid configurations is used. Validity information is obtained by a presampling algorithm, exploring the search space and storing information about the validity of each encountered configuration.
By leveraging the spatial locality of valid candidates found in Section \ref{sec:graph_clusters}, the randomness is reduced and more valid samples are retrieved from the search space. To check validity, the existing VTA compiler is used as a checking function.

\begin{algorithm}[t]
	\footnotesize
	\caption{Algorithm for locality driven search space sampling.}
	\label{alg:presampling}
	\begin{algorithmic}[1]
		\Procedure{Presample}{$n\_{samples}$, \textit{n\_parallel}, $S$}
		\State $N \leftarrow \emptyset$
		%\State $\textit{valid} \leftarrow \text{empty list}$
		%\State $\textit{invalid} \leftarrow \text{empty list}$
		\State $P \leftarrow $ randomly sample \textit{n\_parallel} points from the search space $S$
		\While{$|N| < \textit{n\_samples}$}
		\State $C \leftarrow \emptyset$
		\For{$p \text{ in } P$}
		\State $r_p \leftarrow $ evaluate validity of $p$ with validity check
		\State $N \leftarrow N \cup \{(p, r_p)\}$
		\If{$\text{is-valid}(r_p)$}
		\State $ C \leftarrow C \cup \{ \text{ neighbors of } p \notin C \cup N  \}$
		\Else
		\State $ C \leftarrow C \cup \{\text{ random point  } p \in S \land p \notin C \cup N \}$
		\EndIf
		\EndFor
		\State $P \leftarrow $ randomly sample \textit{n\_parallel} points from the candidates $C$
		\EndWhile
		\State \Return $N$ 
		\EndProcedure
	\end{algorithmic}
\end{algorithm}

Algorithm \ref{alg:presampling} describes how this feature is used to find more valid samples in the search space.
In a first step, a randomly sampled set of configurations $P_0$ from search space $S$ is classified as valid or invalid (line $3$). Each configuration $p \in P$ is added to output set $N$, together with the validity information. If $p$ is valid, locality is exploited by adding all its neighbours to the set of potential candidates for the next evaluation step $C$ (line $10$). Two configurations are neighbours, if their Manhattan distance is 1.
The set of neighbours has a higher chance of containing additional valid configurations.
To maintain exploration, each invalid configuration adds a random new candidate from $S$ to $C$ (lines $11-12$).
After processing all members of $P$, a new set $P$ is sampled from $C$, as shown in line 13. These steps are repeated until the limit of $n_{samples}$ is reached.
The valid and invalid candidates to generate the balanced initialization set for AutoTVM are sampled from $N$.

\subsection{Sample Selection with Distance Maximization}\label{c5-sec:sample_selection}
Presampling, as described in the previous section, generates a set $N$ of configurations and information on their validity. From $N$, the initial measurement epoch $E_0 \subset N$ is selected, such that a desired ratio of valid and invalid configurations is achieved. To do this, $N$ should to be larger than $E_0$, to ensure a degree of diversity in the possible candidates for $E_0$.
A result of presampling algorithm is that $N$ can be defined as $N = N^{valid} \cup N^{invalid}$. Therefore, $E_0$ can be constructed from the configurations $E_0^{valid} \subset N^{valid}$ and $E_0^{invalid} \subset N^{invalid}$ as $E_0 = E_0^{valid} \cup E_0^{invalid}$.
The selection from $E_0^{valid}$ and $E_0^{invalid}$ happens individually, maximizing the Manhattan distance between all selected candidates.

\subsection{Biased Simulated Annealing}\label{c5-sec:bsa}
Since the validity of a search space subset is already known at runtime, this information can be used to bias the simulated annealing. In experiments, the value of performance model predictions were mostly in the interval $[-7,7]$. Known invalids are set to $-10^6$, removing them from the top-$n$ ranking. Known valids are handled with more care. Since we do not know their performance, a place in the top-$n$ cannot be guaranteed. Therefore, we bias known valids with a value of $+1$. This boosts their chance to place in the top-$n$ ranking without overshadowing potentially better implementations. 
\section{Experiments}\label{c5-sec:results}

The previous section presented multiple changes to the auto-tuning process, with the goal to improve the robustness and reduce the convergence time. Mainly the initialization of the model was of concern, as this proved to be a source of instability during the search. The following sections will evaluate the modifications and compare them to the baseline AutoTVM. The comparison will consider both trained and untrained models.

\subsection{Experimental Setup}
All experiments in this section use pre-recorded ground truth data from the VTA hardware, the same as in the search space analysis 
evaluation in Section \ref{c5-sec:searchspaceanalysis}. They act as proxys to real hardware
measurements. This reduces the time to evaluate individual candidates and enables
repeated experiments for robustness evaluation. Details on the operators in the working data set 
are in Table \ref{c5-tab:workload_props}.

Before the actual tuning process begins, the presampling algorithm from section \ref{c5-sec:presampling} explores the search space and gathers the data to replace random initialization of the first measure epoch $E_0$ with a set of configurations created by the process from section \ref{c5-sec:sample_selection}. From there, the existing flow of AutoTVM is used: measurements in epoch $E_i$ are used to train the performance model. This performance model is then used by the simulated annealing algorithm to select the configuration for next measure epoch $E_{i+1}$. This new batch of configurations is then evaluated on hardware and the cycles repeats. The original AutoTVM implementation has been altered in two locations: 1) The simulated annealing now uses the bias information from section \ref{c5-sec:bsa} and 2) hardware evaluations are replaced with querying the ground truth data gathered once for the full search space. All experiments use an epoch size of 50 measurements, or trials, until a total of 750 evaluations on hardware are performed.
In all experiments with the extended auto-tuning, the presampling set $N \subset S$ of each search space $S$ is built form $min(1000, |S|)$ configurations. Model initialization is then performed as described in Section \ref{c5:methods}: $E_0 = E^{valid} \cup E^{invalid}$, with $E^{valid} = min(|N^{valid}|, 25)$ and $E^{invalid} = min(|N^{invalid}|, 50 - |N^{valid}|)$.

\subsection{Evaluation against Baseline}
\begin{figure}[t]
	\centering
	\includegraphics[width=\linewidth]{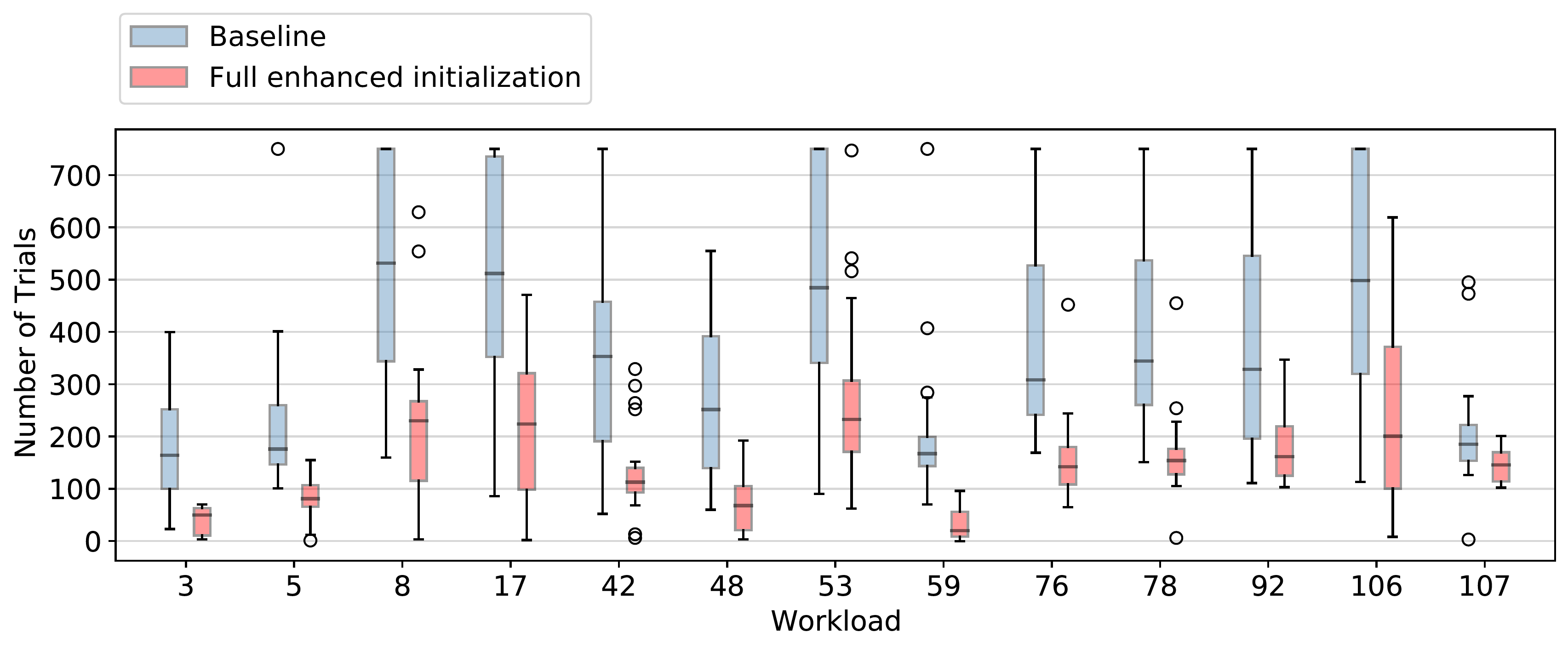}
	% \resizebox*{\textwidth}{!}{%
	% {\huge
	% \input{img/num_trials_.pgf}
	% }
	% }
	\caption[Number of trials: Baseline and full method]{Convergence and robustness of the novel method  (Full enhanced initialization) and the AutoTVM baseline over 20 tuning runs.}
	\label{c5-fig:num_trials_base}
\end{figure}

First,  the effects of the improved model are investigated. Runs with randomly initialized search and an unmodified AutoTVM serve as a baseline.
The results for the individual runs are in Figure \ref{c5-fig:num_trials_base}.
The improved methods outperform the baseline for every workload, with a mean of only $0.416\times$ as many trials for convergence. Two results stand out: layer 59 is remarkably better ($0.111\times $) than the mean and layer 107 shows weaker improvements ($0.735\times $). These are also the workloads with lowest (layer 59) and the highest (layer 107) numbers of valids  in the search space. For layer 59, if there are very few valids, finding any of these has much higher chance of being the best one. Layer 107 on the other hand, has the highest number of valids candidates, meaning that finding valids alone is not sufficient for good performance, making the search process harder. Nonetheless, our method improved the convergence time by $1.36\times$.
%
%Similar results are achieved for the robustness of the search, presented in table \ref{c5-tab:relative_iqr}.
Across all experiments, the mean IQR is only $0.424\times$ of the baseline. In some layers, like 3, 5, 8, 42, 48 or 92, IRQ measure was improved by factors of three to six.

All the median values of the presented method are consistently below 300 trials, with only layers 8, 17 and 53 requiring more than 200 trials. 
%A first overview shows improved robustness and convergence time for all workloads. 
Further, the robustness of the proposed method roughly follows the baseline, but always improves upon it.
Meaning that a layer with narrow IQR in the baseline, will also have a narrow IQR with the improved initialization and a wide IQR in the baseline leads to a wider IQR in the new method. While the overall performance improved, some runs still produced outlier performing significantly worse than the rest, for example layers 8, 42 or 53.
Interesting is layer 42, where the IQR with presampling is narrow, but several positive and negative outliers exist. The baseline, on the other hand spreads from 50 to 750 trials, with IQR between 200 and 450. This hints towards a workload, that is rather difficult for the performance model to reason about and where the initial dataset has a large impact on the rest of the search. This is the complete opposite of layers 3, 5 or 106, where the spread is narrowed considerably and outliers are few or non-existent. 
The high robustness of 59 can again be explained with extremely low number of valid configuration in the search space. Often the best candidate was in the initial training batch.

\subsection{Evaluation against Trained Performance Model}

\begin{figure}[t]
	\centering
	\includegraphics[width=\linewidth]{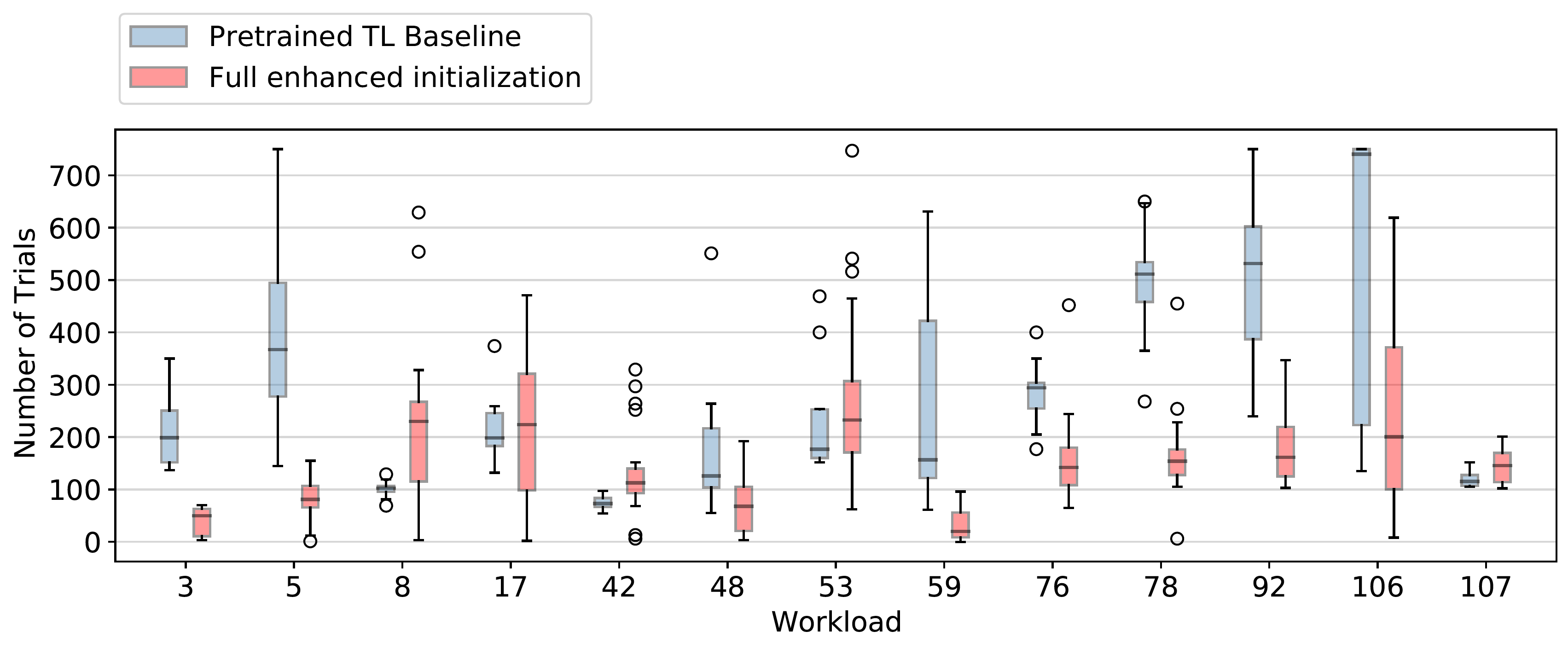}
	% \resizebox*{\textwidth}{!}{%
	% \input{img/num_trials_nodict.pgf}
	% }
	\caption[Number of trials: pretrained Baseline]{ Convergence and robustness for each workload over 20 runs this work (Full enhanced initialization) and 10 runs for the transfer-learning solution (TL Baseline).}
	\label{c5-fig:num_trials_curve2}
\end{figure}
In the previous experiments, the AutoTVM model used random sampling to start the search process and used an uninitialized model, whereas the presampling method already provided validity information for the start of the tuning process. This showed to significantly search performance and robustness.
In the following experiments AutoTVM with a performance prediction model trained on 30000 random samples from all workloads in Table \ref{c5-tab:workload_props} is compared against the presampling based tuning. Since the model only learned about other workloads, this feature is called \textit{transfer learning}.

The experiments with transfer learning will use the context relation features, as they provide the ability to learn across different operators \cite{AutoTVM:10.5555/3327144.3327258}. 
From the fully evaluated search spaces, 30000 configurations were randomly sampled as training data (but never from the workload under evaluation). From these, ca. 63.04\% could be used to train the model. The other samples proved to be invalid configurations during feature extraction phase and thus were removed from the training set.
Although the context relation features are designed to produce feature vectors compatible across workloads, the vectors of some workloads still required zero padding to fit the input length. Due to the large overheads during transfer learning, experiments were only repeated 10 times, instead of 20.

Figure \ref{c5-fig:num_trials_curve2} compares the convergence time of this work and the trained tuner. Overall, the pretrained model is only needs 90.5\% of the trials to find the best result, compared to the AutoTVM baseline. However, in almost half of the experiments, the tuning performance decreased. Five workloads (3, 5, 78, 92, 106) perform worse than the random baseline, layer 5 needs $2.085\times$ as many trials in the median. On the other hand, layers 8 or 42 performed significantly better than the presampling based method by factors of $2.24\times$ and $1.54\times$. 
For the robustness, transfer learning improves the baseline by a factor of $1.107\times$, but has strong variations in both directions. Layers 8 and 42, achieve an exceptional robustness, with the IQR being only 2.7\% and 6.5\% of the baseline. For layer 59, learning from other workloads is not successful, as IQR increases by a factor of $5.386\times$, although the convergence is slightly better.

Overall, the pretrained model consistently requires over 200 samples for median convergence, with the exceptions of layers 42, 53 barely and 107. Layers 8, 42, 53 and 107 show high robustness and no outliers, while outperforming the pre-samping method. However, workloads 5, 78, 92 or 106 are the complete opposite of this. They need over 300, the latter three over 500 samples to converge to the best solution. They also have a low robustness with the outer quartiles reaching up to the sampling limit.
While these results improve upon the random baseline, our presampling based method outperforms the transfer learning approach in more than half the cases and never drops below baseline performance.

\subsection{Tuning Overhead}\label{sec:time}
The previous sections compared the tuning performance by the number of trials, not time-to-solution. This metric makes comparison of different approaches easier and decouples the evaluation from runtime effects. However, these methods also introduce overheads in the overall process. While hardware execution is a major contributor to the overall tuning time, other overheads can increase the absolute runtime. Table \ref{tab:time} lists the per run overheads of the different methods. Performing the validation of individual candidates for model initialization has a significant overhead, while the sample selection overhead is the negligible. For layers 78, 8 and 92 presampling plus sample selection is still 82.6s, 415.4s and 246s faster than transfer learning. However, the presampling has to be performed for every new workload, whereas a once trained model can be reused for different workloads.
Tuning the knob features, which just flatten the configuration into a vector, is faster than the curve features by factors of $8.44\times$, $14.42\times$ and $12.89\times$ for the layers in Table \ref{tab:time}. The data shows that for actual tuning runs, both the pre-tuning and runtime overhead are lower with our method, compared to the transfer-learning solution of TVM. Even when using an already trained model, the presampling overhead is amortized against the much more expensive feature extraction.

\begin{table}[t]
	\caption[Measurement unrelated overhead]{Tuning overheads (in \textit{seconds}) averaged over 10 runs. presampling 
		and sample selection for the work presented in this section. pretrainning for transfer learning with context relation features. Features based on AutoTVM.}
	\centering
	\label{tab:time}
	%        \resizebox*{\textwidth}{!}{
	\begin{tabular}{rrrr|rr}
		\multicolumn{3}{l}{Pre-tuning Overhead}&& \multicolumn{2}{l}{Runtime Overhead} \\
		\hline
		& \multicolumn{1}{l}{Pre-}    & \multicolumn{1}{l}{Sample} & \multicolumn{1}{l}{Pre-} & &\multicolumn{1}{l}{Context-}\\
		\multicolumn{1}{l}{Layer} &\multicolumn{1}{l}{Sample} &\multicolumn{1}{l}{selection} &\multicolumn{1}{l}{training}&\multicolumn{1}{l}{Knob}&\multicolumn{1}{l}{Relation}\\
		\hline
		78 & 416.8 & 13.4 & 512.8 &77.8 &656.8  \\
		8  & 375.5 & 13.0 & 839.9 &77.5 &1118.7 \\
		92 & 396.0 & 11.7 & 653.7 &88.4 &1139.7 \\
		\hline
		
	\end{tabular}   
\end{table}

\section{Discussion and Related Work}\label{sec:discussion}

Prior art on auto-tuning is mostly focussed on improvements of the underlying ML-model and the search algorithm to improve result quality. In \textsc{chameleon}\cite{Chameleon:48873}, SA and boosted trees are replaced with a reinforcement-learning agent. Further, sampling time was reduced by using only a single hardware measurement for similar configurations. Likewise, AutoHalide \cite{adams2019learning} and Ansor \cite{Ansor:258858} further evolve the search processes and underlying models. However, the influence of invalid configurations is not discussed, as they are mostly focussed on x86 and GPU architectures, where this issues is much less dominant. Another approach is followed by \textsc{Telamon} \cite{Telamon:hal-01655602}, which avoids invalid configurations by relying on a constraint based, manually crafted hardware model. The model predicts the upper performance bound, while avoiding the construction of invalid configurations. Similarly, DORY \cite{Dory} builds a constraint model of the PULP hardware's memory hierarchy. These hand-crafted approaches deliver good performance, however come at high engineering cost, requiring deep hardware knowledge. 

Auto-tuning and performance optimization in general are a tradeoff between data cost, engineering cost, measurement cost. For general purpose systems with an abundance of measurement data, like GPUs in a cloud environment, fully data driven approaches can generate impressive results quickly, at the cost of millions of training samples \cite{10.1145/3445814.3446762}, while excluding all invalid candidates from search and training. This is not always feasible for specialized devices, like embedded systems where  sufficient training data is not readily available. Approaches with higher engineering cost can be a hurdle in some contexts, like hardware development. While feedback driven approaches are more portable, the reduction of hardware measurements is the key to reduce optimization times. In this context, the presampling method for search spaces with a large share of invalid configurations demonstrated significant performance improvements over the baseline. The experiments with the curated initial data set and the transfer learning showed how important the composition of the training set is to the auto-tuning performance. In the transfer-learning scenario, the training set only contains information about a similar but not identical problems. The individuality of each workload with respect to valid and invalid configurations is not yet learned at this point. A possible explanation for this is feature complexity. The context relation features used for transfer learning provides a more detailed view of a candidate's execution, at the cost of significantly larger feature vectors, compared to the cheaper knob features. And while the knobs are a simpler representation, they accurately capture both validity and performance information. The context relation features lose this specificity to gain portability.
Improving the initialization point is not a novel concept, per se. Algorithms like \texttt{kmeans++}~\cite{10.5555/1283383.1283494} already demonstrated the benefits of starting point optimizations.

\section{Conclusion}

While auto-tuning itself is already an established approach to DNN optimization, dedicated accelerators can add another dimension to this problem.

Insights into the solution distribution and behaviour of the performance prediction model in the light of a dominant subspace of invalid configurations form the base for a new initialization process of the auto-tuner. First, the search space is sampled for valid and invalid configurations, without considering performance, yet. Since clusters of valid configurations were found in the search spaces, the presampling algorithm utilizes this to exploit these local pockets of possible solutions. Otherwise, the space is explored randomly.
From the information of valid and invalid subsets, a curated training set is created.

Experiments show that this method consistently outperforms the baseline with random initialization and is more robust than tuning with a pretrained model. The median time to find the best solution is reduced across all workloads and the difference between individual runs for the same workload are minimized. Experiments with transfer learning showed that both, quality and quantity of training data need consideration.

For real world applications, this new process can bring two advantages. The first is more confidence in the selected tuning duration. In practice, we do not know in advance if the solution has been found, thus the search usually runs longer than necessary. With the improved robustness, tuning can end sooner and give greater confidence in the result.

Second, the process only relies on data for a single workload and is not dependent on existing training data. This is helpful for non-commodity hardware and HW/SW-Codesign.

What remains to be explored is the generalization of this method. How well it translates to other hardware architectures is highly dependent on both the architecture itself and code generation. However, since the primary insight is that a more nuanced initialization of the prediction model yields better tuning performance, the method is agnostic to the source of the invalid configuration. If the performance prediction model can interpret validity from the presented features, a generalization should be possible.

\bibliographystyle{acm}
\bibliography{dissertation}

\end{document}